\let\oldvec\vec  
\documentclass{llncs}
\let\vec\oldvec  

\usepackage{amsmath}
\usepackage{booktabs}
\usepackage{nicefrac}
\usepackage{graphicx}
\usepackage{siunitx}

\newcommand*{\etal}{et~al.}
\newcommand*{\myFig}{Fig.}

\begin{document}

\title{Multi-dimensional Gated Recurrent Units for Automated Anatomical Landmark Localization}
\author{Simon~Andermatt\textsuperscript{*,1} \and Simon~Pezold\textsuperscript{*,1} \and Michael~Amann\textsuperscript{2,3,4} \and Philippe~C.~Cattin\textsuperscript{1}}
\institute{
\textsuperscript{1}Department of Biomedical Engineering, University of Basel, Allschwil, Switzerland\\
\textsuperscript{2}Department of Neurology, University Hospital Basel, Basel, Switzerland\\
\textsuperscript{3}Department of Radiology, University Hospital Basel, Basel, Switzerland\\
\textsuperscript{4}MIAC AG, Basel, Switzerland\\
\textsuperscript{*}S.\,Andermatt and S.\,Pezold contributed equally.
}
\maketitle

\begin{abstract}
We present an automated method for localizing an anatomical landmark in three-dimensional medical images. The method combines two recurrent neural networks in a coarse-to-fine approach: The first network determines a candidate neighborhood by analyzing the complete given image volume. The second network localizes the actual landmark precisely and accurately in the candidate neighborhood. Both networks take advantage of multi-dimensional gated recurrent units in their main layers, which allow for high model complexity with a comparatively small set of parameters.
We localize the medullopontine sulcus in 3D magnetic resonance images of the head and neck. We show that the proposed approach outperforms similar localization techniques both in terms of mean distance in millimeters and voxels w.r.t. manual labelings of the data. With a mean localization error of $\SI{1.7}{\milli\meter}$, the proposed approach performs on par with neurological experts, as we demonstrate in an interrater comparison.

\end{abstract}
\section{Introduction\label{sec:intro}}

Localizing anatomical landmarks is a common task in many medical applications. Finding matching anatomical points in images may be necessary for seeding a segmentation algorithm, for registration problems, or for providing points of reference for quantitative measurements. 
Although finding landmarks in volumetric images is error-prone and time-consuming, the task is often still carried out manually. Using a fully automated approach mitigates the inter and intra-rater variability through an objective and efficient process without manual interference. Therefore, many automated localization methods have been proposed, with varying degrees of robustness, reliability, and generalization potential.
Some of the methods, such as Bhanu Prakash {\etal}~\cite{bhanu_prakash_rapid_2006} or Elattar {\etal}~\cite{elattar_automatic_2016}, use very basic image processing techniques, but many others rely on concepts from machine learning: for example, for localizing landmarks in the brain, Guerrero {\etal}~\cite{guerrero_laplacian_2011} use manifold learning and O'Neil {\etal}~\cite{oneil_cross-modality_2016} use random forests; for cardiac landmark localization, Karavides {\etal}~\cite{karavides_database_2010} use Adaboost and Lu and Jolly~\cite{lu_discriminative_2012} use probabilistic boosting trees; Xue {\etal}~\cite{xue_automatic_2015} use boosting for localizing landmarks on the knee joint. For a recent overview, also see Zhou {\etal}~\cite{zhou_discriminative_2014}. 

In recent years, ground-breaking advancements using neural networks have been achieved in various domains, allowing for automatic learning of discriminative features for the problem at hand and avoiding the need for manually designed (often called handcrafted) features. Consequently, these techniques have also found their way into landmark localization. Examples are Zheng {\etal}~\cite{zheng_3d_2015}, who use two neural networks successively to localize the carotid bifurcation in 3D CT images, Ghesu {\etal}~\cite{ghesu_artificial_2016}, who propose a so-called artificial agent for localizing various anatomical landmarks in 2D and 3D images of different modalities, and Yang {\etal}~\cite{yang_automated_2015}, who apply convolutional neural networks for landmark localization on the femur in MR images.

Existing approaches based on convolutional neural networks (CNNs) are capable of detecting very delicate structure, yet are limited to the local neighborhood of the filters used in each layer of the network. Using a recurrent neural network (RNN) for this task allows for flexible feature relationships of varying length and scale. This is especially useful given a localization task, where the surrounding tissues structure can take a number of different shapes and sizes. Tackling volumetric data with RNNs for \emph{segmentation} has been recently demonstrated by Andermatt {\etal}~\cite{andermatt_multi-dimensional_2016} with multi-dimensional gated recurrent units (MD-GRUs). To our knowledge, neither multi-dimensional RNN nor MD-GRUs have been applied to the task of \emph{landmark localization} so far.

In this paper, we propose to apply MD-GRUs in a two-stage approach to the task of anatomical landmark localization. In the first stage, the anatomical region of interest is roughly located in the given image volume. We then determine the actual landmark coordinate in a subvolume in the second stage. We apply the proposed method to 3D MR images of the head and neck, in which we locate the medullopontine sulcus, and compare the found coordinates to those of manual labels. 
Our results from an interrater comparison suggest that the proposed method cannot be distinguished from a clinical expert.

\section{Methods}

\begin{figure}[t]
\centering
\includegraphics[width=0.8\textwidth]{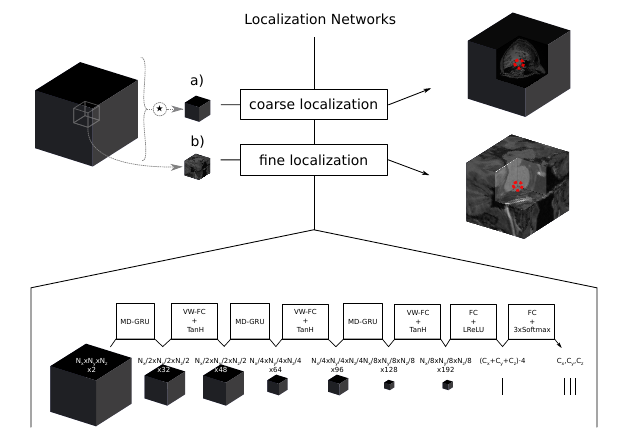}
\caption{Localization network. a) Coarse approximation of landmark coordinates in subsampled low resolution representation of full data. b) Fine approximation of landmark coordinates in extracted window around detected coarse location in a second localization network. Both networks use the architecture depicted at the bottom.}
\label{fig:schematic}
\end{figure}

For the accurate localization of landmarks, we propose to use two separate localization networks of similar structure, to both accelerate the process and allow for a decently complex network. Both localization networks work on the same number of voxels -- in our case we fixed it to $64^3$~voxels -- and find the coordinate in said volume which lies closest to the true landmark. The first network is provided data subsampled to such a degree, that the full original volume can be represented inside of it. The network will then approximate a location, which will in turn be used to sample a subvolume at the original resolution from the image data around the found location. In our case, the first network is provided with 4-fold subsampled data and the second processes data at the original resolution, centered at the location which was found by the first network.
\subsubsection{Subsampling MD-GRU Layer}
We propose to adapt the MD-GRU layer \cite{andermatt_multi-dimensional_2016}, which was introduced to handle segmentation problems, to the application of landmark localization. In order to do so, we implement the ability to subsample at each MD-GRU layer and hence at each convolutional gated recurrent unit (C-GRU) which it consists of. This effectively reduces the spatial problem size, allowing a multi-resolution processing approach. We adjust the original C-GRU equations as follows:
\begin{alignat}{2}
\label{strided_conv}&f^j(t,\alpha,\beta) = \sum_i^I x_t^i \star \alpha^{i,j}+\beta^j , \hspace{2em}&&g^j(t,\alpha) = \sum_k^J h_{t-1}^k*\alpha^{k,j},\\
\label{reset_update_gate} &r_t^j = \sigma(f^j(t,w_r,b_r) + g^j(t,u_r)), \hspace{2em}&&z_t^j = \sigma(f^j(t,w_z,b_z)+g^j(t,u_z)),\\
\label{proposal_and_state} &\tilde{h}^j_t = \phi(f^j(t,w,b)+r_t^j \odot g^j(t,u)), \hspace{2em}&&h^j_{t} = z_t^j\odot h^j_{t-1} + (1-z_t^j)\odot\tilde{h}^j_{t},
\end{alignat}
where $x_t^\cdot$, $h_t^\cdot$ denote the input and state of the C-GRU at time $t$, and $i$, $j$, $k$ denote the respective channels. The operator $\odot$ denotes elementwise multiplication, as in \cite{andermatt_multi-dimensional_2016}. Variables $u$, $w$, and $b$ are trainable weights. 
We call $\tilde{h}$ in Eq.~\eqref{proposal_and_state} the proposal and $r$ and $z$  in Eqs.~(\ref{reset_update_gate}) the reset and update gate.

We accomplish subsampling by introducing strided convolutions, which are denoted as $\star$ in Eq.~(\ref{strided_conv}). The size of the state as well as of all the gates and the proposal will be reduced by the factor of the chosen stride $S$ per spatial dimension. Each C-GRUs' output is then subjected to one-dimensional average pooling, compressing the time dimension by stride $S$. The sum of all $d$ compressed C-GRU results $\hat{h}$ yields the MD-GRU output $H$:
\begin{align}
\label{last_strided_convolution} H^j=\sum_{d}\hat{h}^j,\hspace{2em}  \hat{h}^j_{t'} = \frac{1}{S}\sum\limits_{s=0}^{S-1} h^j_{St'+s}.
\end{align}
\subsubsection{Localization Network}\label{locnet}
At the core, we use the same localization network for all experiments. We use three subsequent compositions of a subsampling MD-GRU layer, a voxelwise fully connected layer, and a tanh activation function. The subsampling MD-GRU layers are provided with 32, 64, and 128~channels, respectively. All of them use strides of 2 along spatial dimensions, the volume is hence subsampled 8-fold at each composition. We use DropConnect \cite{wan2013regularization} with a drop rate of 0.5 on the input convolution filters of both gates $r^j,\,z^j$ and the proposal $\tilde{h}$. The voxelwise fully connected layers are realized through convolution layers with spatial filters of $1^3$, with 48, 96, and 192~channels each. 

The resulting subvolume is of size $\nicefrac{N_x}{8} \times \nicefrac{N_y}{8} \times \nicefrac{N_z}{8}$, given the input shape was $(N_x \times N_y \times N_z)$. The subvolume is reshaped into a vector, in which we process each coordinate by two fully connected layers of $(C_x+C_y+C_z) \cdot 4$ and $(C_x+C_y+C_z)$ layers, which are connected through a leaky rectifying unit defined as $\text{lrelu}(x) = \max\{0.01\,x,\:x\}$. The resulting vector is split into three separate vectors of sizes $C_x$, $C_y$, and $C_z$, where $C_.$ gives the number of possible coordinate positions along the respective dimension. These are then fed into individual softmax activation functions to estimate the probabilities for each coordinate in each vector. We use the sum of all cross entropy losses as loss function for the entire network. Figure~\ref{fig:schematic} shows an overview of the network architecture.

\subsubsection{Subsampling}
In the first stage, we use a strided convolution on the input to match the localization networks input resolution. We pad the input, such that the shape of the volume is a multiple of the required shape for the localization network.
In our case, we padded the data to $256^3$ and used strides $S$ of 4 with a filter size of $S \cdot 2+1$ and 16~channels for the convolution layer.

\subsubsection{Superresolution}
Our method, as explained so far, is restricted to voxel coordinates, since we estimate with our method discrete instead of continuous coordinates. In the following, we explain two extensions to our idea to yield superresolution results. 

The first extension takes advantage of the coordinate resolution-independent formulation in the \emph{Localization Network} paragraph above.
Instead of estimating as many classes for each of the three coordinates as there are voxels in the respective dimension in the volume, we estimate $n$ times the amount. This allows us to estimate values which are $\nicefrac{1}{n}$ voxels apart and hence allow for a more fine-grained localization. In our experiments, we use $n=4$ resulting in 256 classes.

Our second idea exploits neighborhood information in our coordinate probability vectors by fitting a parabola to the largest probability and its two neighbors per coordinate. The maxima of these functions can then be interpreted as our coordinate location. This allows for an even finer localization, but is based and hence limited on the chosen number of coordinate probabilities.%

\subsubsection{Optimization}
\begin{figure}[t]
\centering
\includegraphics[width=\textwidth]{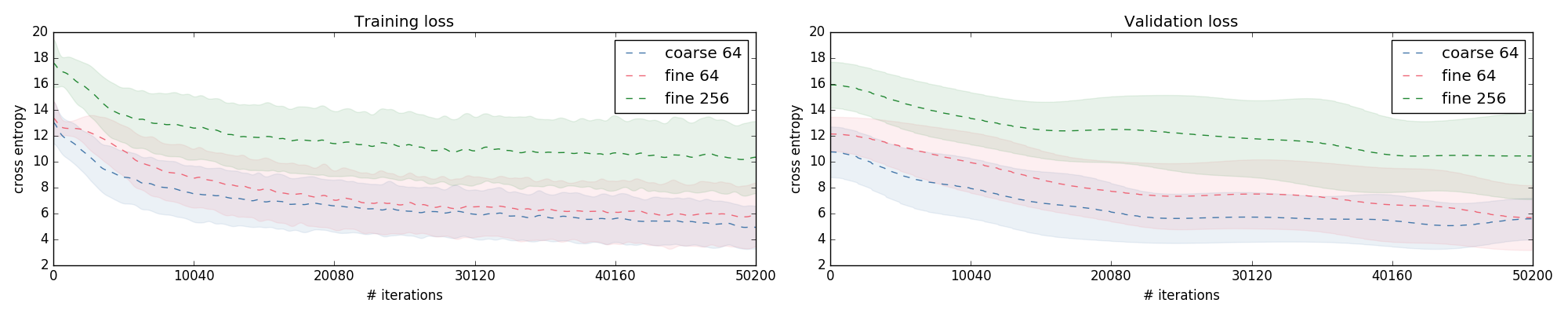}
\caption{Cross entropy loss. Mean $\pm$ one standard deviation on training and validation set for the 3 trained networks, smoothed using a gaussian for visualization.}
\label{fig:training}
\end{figure}

We trained each localization network together with their subsampling addition individually. All networks were trained for a total of 50~epochs, where one epoch comprised one random sample from each training subject, which led to a total of 50\,200~iterations. We used AdaDelta \cite{zeiler_adadelta:_2012} with a learning rate of 0.001. We initialized all weights of the convolutions with the method of Glorot and Bengio \cite{glorot2010understanding}, the biases with zero and the fully connected layers at the end of the localization network with random values from $[-\nicefrac{\sqrt{3}}{N_i},+\nicefrac{\sqrt{3}}{N_i}]$, where $N_i$ is the number of input units. For the first network, we sampled from the center of the padded volume with a random offset in the range of $[-100,100]$ voxels per coordinate; for the second network, we just required that the training landmark was within the volume. The training loss is visualized in Fig.~\ref{fig:training}.

For preprocessing, we apply a high-pass filter on the input, the results of which we use together with the original data as input to our networks. Additionally, we normalize to zero mean and a standard deviation of one for each of the input volumes. Apart from this, no preprocessing is required.  
\section{Results\label{sec:results}}

\begin{figure}[t]
\centering
\includegraphics[height=.155\textwidth, angle=90]{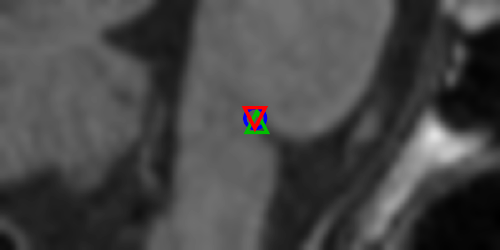}
\includegraphics[height=.155\textwidth, angle=90]{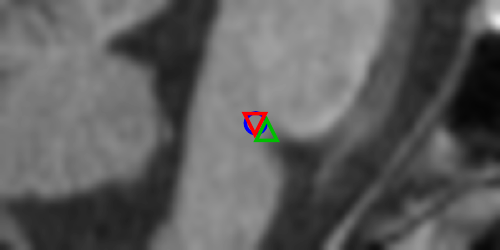}
\includegraphics[height=.155\textwidth, angle=90]{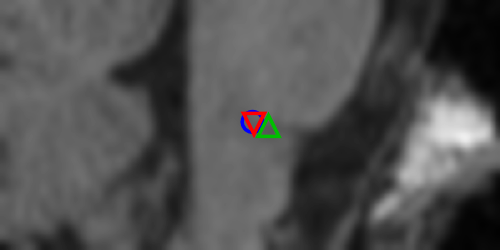}
\hfill
\includegraphics[height=.155\textwidth, angle=90]{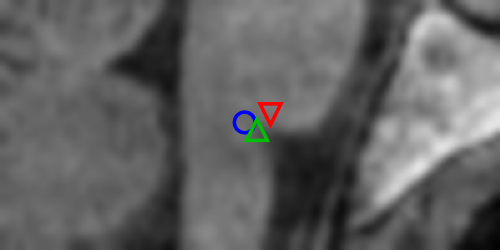}
\includegraphics[height=.155\textwidth, angle=90]{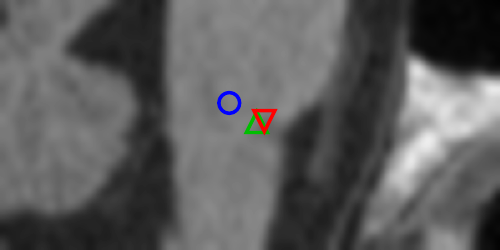}
\includegraphics[height=.155\textwidth, angle=90]{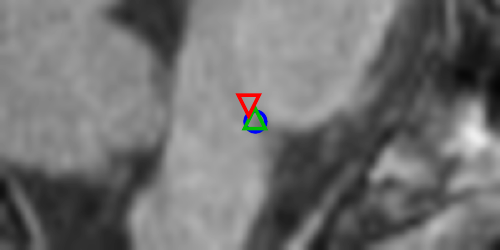}

\includegraphics[width=.155\textwidth, angle=0]{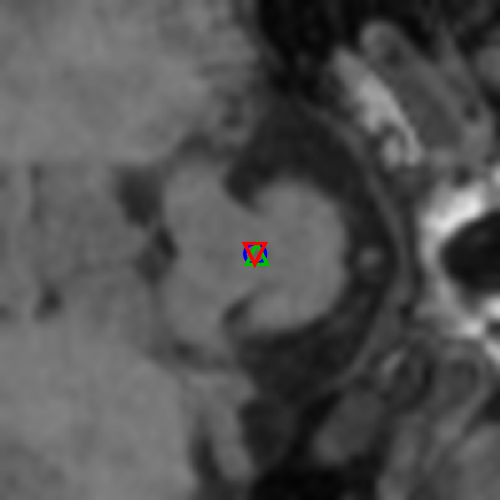}
\includegraphics[width=.155\textwidth, angle=0]{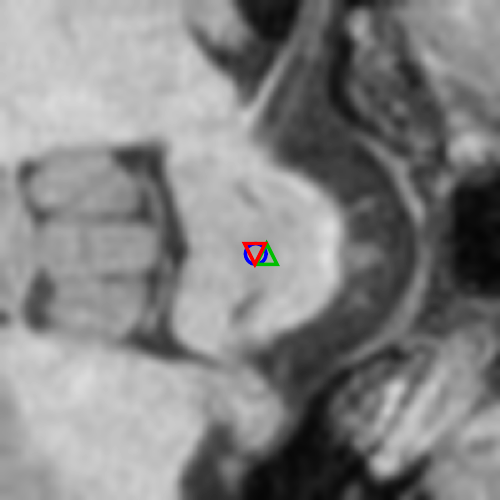}
\includegraphics[width=.155\textwidth, angle=0]{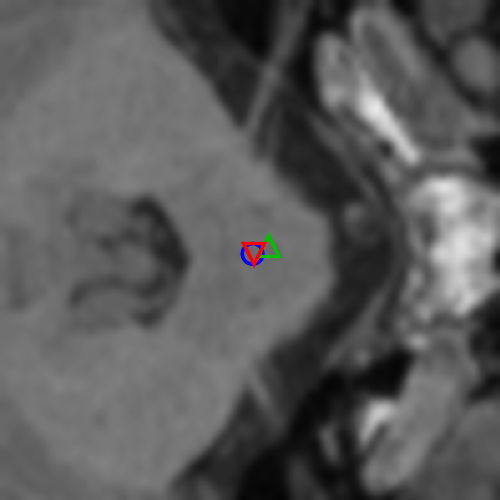}
\hfill
\includegraphics[width=.155\textwidth, angle=0]{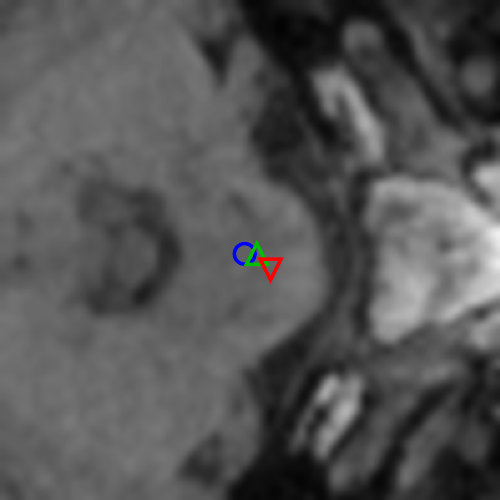}
\includegraphics[width=.155\textwidth, angle=0]{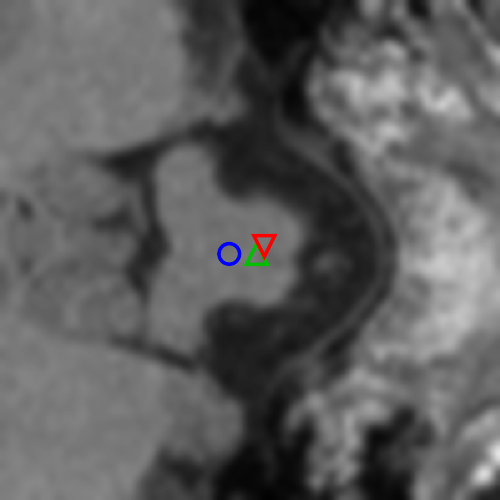}
\includegraphics[width=.155\textwidth, angle=0]{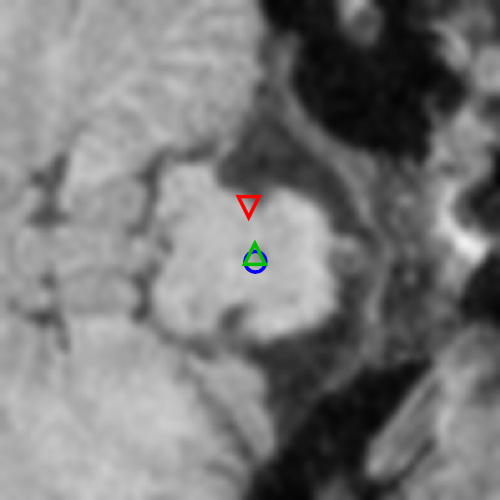}
\caption{Localization results for rater~1 \emph{(red {\scriptsize $\bigtriangledown$})}, rater~2 \emph{(green {\scriptsize $\bigtriangleup$})}, and proposed method \emph{(blue {\large $\circ$})}. Shown are the best three \emph{(left)} and worst three \emph{(right)} localizations of the proposed method wrt. rater~1, both in sagittal \emph{(top)} and transverse \emph{(bottom)} view.}
\label{fig:manual-vs-auto}
\end{figure}

\begin{table}[t]
\caption{\label{tab:results-coarse-vs-fine}Localization accuracy and precision. a)~Localization error on the test set when using only the first network (\emph{top row}) and both networks with a varying number of coordinate classes, with or without parabola fitting (\emph{bottom row:} proposed combination); b)~localization error on the test set in comparison to two human raters; c)~localization errors reported in the literature.} 

\begin{tabular}[t]{lllll}
\toprule 
a) &  & \multicolumn{3}{c}{Error [mm]}\tabularnewline
\cmidrule{3-5}
 &  & Median & Mean & Std.\tabularnewline
\midrule
\midrule 
Coarse localization &  & 4.83 & 5.02 & 2.22\tabularnewline
Fine, 64 classes &  & 1.74 & 1.97 & 1.02\tabularnewline
Fine+parab., 64 cl. &  & 1.77 & 1.89 & \textbf{0.98}\tabularnewline
Fine, 256 classes &  & 1.47 & 1.72 & 1.03\tabularnewline
Fine+parab., 256 cl. &  & \textbf{1.40} & \textbf{1.69} & 1.02\tabularnewline
\bottomrule
\end{tabular}
\hfill
\begin{tabular}[t]{lllll}
\toprule 
b) &  & \multicolumn{3}{c}{Error [mm]}\tabularnewline
\cmidrule{3-5}
 &  & Median & Mean & Std.\tabularnewline
\midrule
\midrule 
Rater\,1 vs. rater\,2 &  & \textbf{1.39} & \textbf{1.59} & 0.98\tabularnewline
Proposed vs. rater\,1 &  & 1.40 & 1.69 & 1.02\tabularnewline
Proposed vs. rater\,2 &  & 1.65 & 1.73 & \textbf{0.87}\tabularnewline
Proposed vs. both &  & 1.50 & 1.71 & 0.95\tabularnewline
\bottomrule
\end{tabular}

\begin{tabular}[t]{lllllllllll}
\toprule 
c) &  & \multicolumn{3}{c}{Error [mm]}\tabularnewline
\cmidrule{3-5}
Method &  & Median & Mean & Std. &  & Voxel size [$\SI{}{\milli\meter\cubed}$] &  & Target landmark\tabularnewline
\midrule
\midrule 
Proposed &  & 1.50 & 1.71 & 0.95 &  & $ 1.00 \times 1.00 \times 1.00$ &  & medullopontine sulcus\tabularnewline
Zheng {\etal}~\cite{zheng_3d_2015} &  & 1.21 & 2.64 & 4.98 &  & $0.46 \times 0.46 \times 0.50$ &  & carotid bifurcation\tabularnewline
Ghesu {\etal}~\cite{ghesu_artificial_2016} &  & 0.8 & 1.8 & 2.9 &  & $ 1.00 \times 1.00 \times 1.00$ &  & carotid bifurcation\tabularnewline
Yang {\etal}~\cite{yang_automated_2015} &  & --- & 4.13 & 1.70 &  & $ 0.37 \times 0.37 \times 0.70$ &  & femoral medial distal point\tabularnewline
Xue {\etal}~\cite{xue_automatic_2015} &  & --- & 1.41 & 0.91 &  & $ 0.3 \times 0.3 \times [0.6, 3]$ &  & knee joint (23 landmarks)\tabularnewline
Guerrero {\etal}~\cite{guerrero_laplacian_2011} &  & --- & 0.45 & 0.22 &  & \multicolumn{1}{c}{---} &  & anterior commissure\tabularnewline
\bottomrule
\end{tabular}
\label{tab:results}
\end{table}

To evaluate the proposed approach, we located the medullopontine sulcus, a distinct cavity in the brainstem, in MR images of the head and neck (see \myFig~\ref{fig:manual-vs-auto}). Images were acquired with a T1-weighted MPRAGE sequence, having a resolution of $\SI{1}{\milli\metre\cubed}$ and a size between $160 \times 240 \times \SI{256}{voxels}$ and $192 \times 256 \times \SI{256}{voxels}$. Altogether, we had $1218$~images of $265$~subjects, with a median number of $5$~images per subject (minimum: $1$, maximum: $8$), which we randomly assigned to a training set ($1004$~images of $213$~subjects), a validation set ($114$~images of $26$~subjects), and a test set ($100$~images of $26$~subjects), making sure that all images of each subject were assigned to the same set.

For training and evaluation of the localization, we used manual labels of the landmark. These labels were provided by clinical expert raters who placed them on a graphical user interface enabling them to zoom in and out of the imaged volumes as necessary. To allow for interrater comparisons, we had two raters place the landmark in all images of the test set.

Training 50 epochs for the coarse and fine networks took around 41 and 34 hours, respectively. Testing, on the other hand, requires less than 2 seconds for either network, resulting in a total of around 3--4 seconds for localization. Using our extension of estimating 256 class probabilities instead of 64 per coordinate requires only 2.5 hours more training time and took around 2.5 seconds per volume for testing, which results in around 4 seconds in total for localization.

Figure~\ref{fig:manual-vs-auto} shows our three best and worst localization results. Note that our largest error (rightmost column in Fig.~\ref{fig:manual-vs-auto}) is actually produced by a mislabeling of a clinical expert, as can be seen by the off-center position of the red marker.

Table~\ref{tab:results}a shows the localization errors when using only the first network as compared to using both. The second network increases the localization accuracy notably, as does using more coordinate classes and fitting a parabola.

Table~\ref{tab:results}b shows the results from comparing both human raters with the proposed approach. The listed values indicate that our approach almost reaches human performance: comparing our results to those of a human rater produces approximately the same error as two human raters compared to each other.

Table~\ref{tab:results}c shows results for landmark localization reported in the literature. 

\section{Discussion and Conclusion}
Our results, as listed in Table~\ref{tab:results}c, appear competitive: compared to other neural network approaches \cite{ghesu_artificial_2016,yang_automated_2015,zheng_3d_2015}, mean error and standard deviation are better in terms of millimeters and voxels. When comparing to Xue {\etal}~\cite{xue_automatic_2015}, one has to keep in mind their notably higher in-plane resolution. While Guerrero {\etal}~\cite{guerrero_laplacian_2011} achieve higher accuracy and precision, a comparison appears difficult: apart from not stating the voxel size, their method requires images with similar field of view, which cannot be guaranteed in our case, as parts of our images are centered on the neck while others are centered on the head. In any case, caution has to be taken when comparing these results: on the one hand, evaluated anatomical landmarks, imaging modalities, and image resolutions differ. On the other hand, our interrater comparison (recall Table~\ref{tab:results}b) suggests that there is a lower bound for the achievable accuracy, which might be well above a given image resolution and might depend on the particular anatomical landmark. Determining the limit of actually achievable accuracy of our method would require evaluating data with lower interrater variability. The results of Xue {\etal}~\cite{xue_automatic_2015} allow a similar conclusion, in that their method's error is similar to the error from their interrater comparison, as well. Unfortunately, the other authors do not provide interrater comparisons.

We have shown two ideas that improved our localization results. The combination of both even surpassed the accuracy of each of them applied separately. Considering interrater variability, we are still slightly less accurate than a human rater. We think that this is partly based on the discrete probability distribution and our sampling technique when training the algorithm. We randomly sampled subvolumes using integer coordinates during training since this process does not require interpolation. But this also means that each training sample could only get mapped on a subset of all possible coordinate classes.

\subsubsection{Conclusion}
We have shown that the localization of the medullopontine sulcus is successfully possible using our proposed automated technique, which adapts MD-GRUs to the task of landmark localization. We introduced a number of improvements, which all led to even more accurate results without significantly increasing the training time. Future work will focus on evaluating our localization approach on multiple anatomical landmarks in different imaging modalities.

\bibliographystyle{splncs03}
\bibliography{biblio}
\end{document}